\documentclass[sigconf,nonacm]{acmart}

\setcopyright{none}
\acmConference[ICAAI '26]{The 10th International Conference on Advances in Artificial Intelligence}{September 21--24, 2026}{London, United Kingdom}
\acmYear{2026}

\usepackage{booktabs}
\usepackage{float}
\usepackage{microtype}

\graphicspath{{./}}

\begin{document}

\title[When the Judge Changes, So Does the Measurement]{When the Judge Changes, So Does the Measurement: Auditing LLM-as-Judge Reliability}

\author{Zongyou Yang}
\affiliation{%
  \institution{Dyson School of Design Engineering, Imperial College London}
  \city{London}
  \country{United Kingdom}}
\email{zy2926@ic.ac.uk}

\author{Yinghan Hou}
\affiliation{%
  \institution{Department of Electrical and Electronic Engineering, Imperial College London}
  \city{London}
  \country{United Kingdom}}
\email{yh24@ic.ac.uk}

\author{Xiaokun Yang}
\authornote{Corresponding author.}
\affiliation{%
  \institution{School of Electronic Information, Nanchang Institute of Technology}
  \city{Nanchang}
  \country{China}}
\email{yangxk@bupt.cn}

\renewcommand{\shortauthors}{Z. Yang, Y. Hou, and X. Yang}

\begin{abstract}
An LLM-as-judge score can move even when the candidate responses stay fixed, simply because the evaluator has changed. We treat this evaluator-replacement ambiguity as a measurement-validity problem. Across four judgment datasets, we compare two upgrade paths available in practice: scaling Qwen3 dense judges from 1.7B to 32B parameters and moving across MiniMax M2--M2.7 released APIs. The main pattern is that judge upgrades are not interchangeable: only Qwen3 1.7B$\rightarrow$4B gives a robust adjacent gain, while MiniMax adjacent releases do not. Stronger judges reduce but do not remove position and verbosity bias. Repeated-sample juries add little when errors are correlated. Structured debate can move decisions substantially, but without parser and fallback logs those shifts cannot be attributed to deliberation. We argue that LLM-as-judge reports should include dataset slices, bias probes, error-dependence estimates, and protocol audit trails.
\end{abstract}



\maketitle

\section{Introduction}

LLM-as-judge evaluation is now often used as a measuring instrument for model quality. The difficulty is that the instrument is itself a model. When a system's score changes after replacing the evaluator, the change is ambiguous: it may indicate that the new judge is more capable, that it is biased differently, that it fails on a different slice of the benchmark, or that the evaluation pipeline parsed and aggregated its outputs differently. This ambiguity is not a minor implementation detail. It determines whether an LLM-as-judge score can be interpreted as evidence about the candidate systems at all.

We call this the \emph{evaluator-replacement ambiguity}: when the measured preference outcome changes after replacing the judge, the source of the change is not identifiable from accuracy alone. This paper studies the ambiguity through two observable interventions available in evaluation practice. The first is a parameter-scaling decision, represented by Qwen3 dense judges from 1.7B to 32B parameters \citep{qwen3}. The second is a released-model upgrade path, represented by MiniMax M2--M2.7 APIs evaluated as released \citep{minimax2026m2}. These axes are evidence sources rather than the paper's object of explanation. The MiniMax M2 report documents the released series and its agent-oriented training pipeline, but it does not make our API sequence a controlled ablation; consequently, this paper does not make a causal claim about MiniMax internals.

This formulation leads to three research questions. RQ1 asks whether judge reliability improves similarly along a parameter axis and a released-model upgrade path. If evaluator capability is the dominant factor, both axes should show consistently positive adjacent gains across datasets; if reliability is workload-dependent, gains should vary by dataset and significance should not transfer uniformly. RQ2 asks whether higher aggregate accuracy also reduces known judge biases. Higher-capability judges should be less bias-sensitive, but nonzero flip rates would show that capability does not eliminate measurement artifacts. RQ3 asks whether protocol-level upgrades, such as juries and debate, change reliability beyond single-judge scaling. Jury gains should be limited when error correlation is high, while debate shifts should depend on capability asymmetry and protocol auditability.

The central thesis is that evaluator capability is important but incomplete: the same judge can look different across workloads, bias probes, correlated votes, and protocol implementations.

This paper contributes an auditable measurement framework for LLM-as-judge reliability, separating model capability from workload dependence, bias behavior, correlated voting, and measurement artifacts. First, it presents a two-axis evaluator study that compares parameter scaling with released-model generation while explicitly avoiding causal claims about MiniMax internals. Second, it decomposes reliability into single-judge accuracy, bias probes, correlated jury errors, and debate auditability. Third, it distills a reporting standard: benchmark slices, parse-shared tests, A/B randomization, $\rho$ estimates, and protocol logs.

The study yields three main results. First, judge upgrades are not interchangeable: parameter scaling produces one robust low-end gain, while adjacent MiniMax releases do not. Second, higher accuracy reduces but does not remove bias. Third, aggregation protocols are interpretable only when their assumptions are auditable: jury voting is bounded by correlated errors, and debate shifts cannot be attributed to deliberation without parser logs.

\begin{figure*}[t]
\centering
\includegraphics[width=0.96\textwidth]{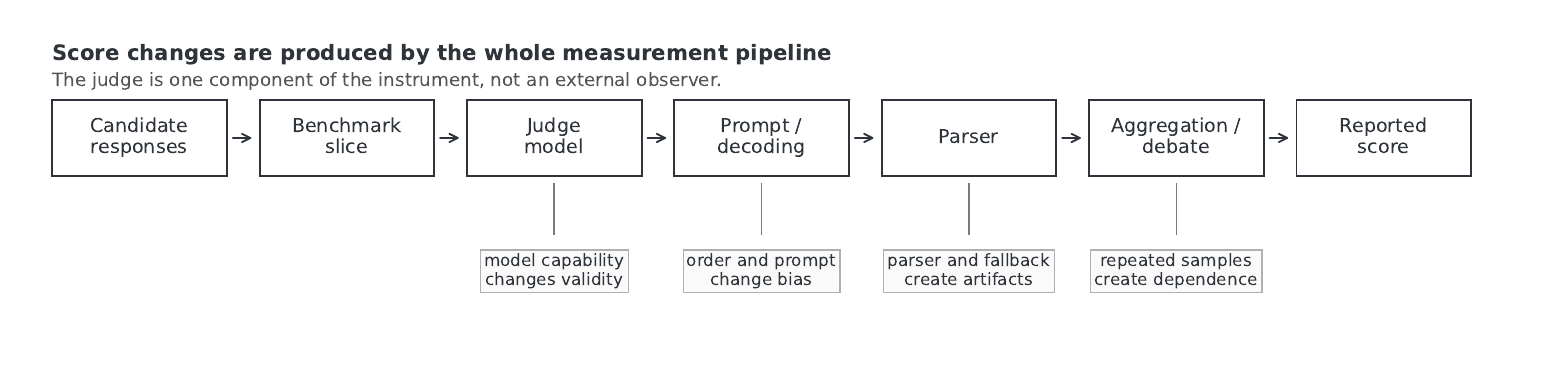}
\caption{LLM-as-judge as a measurement pipeline. A reported score depends not only on candidate responses, but also on the benchmark slice, judge model, prompt and decoding choices, parser, and any aggregation or debate protocol.}
\Description{A pipeline diagram showing candidate responses, benchmark slice, judge model, prompt and decoding, parser, aggregation or debate, and reported score, with risk labels for validity, bias, dependence, and parser artifacts.}
\label{fig:pipeline}
\end{figure*}

\section{Related Work}

\textbf{LLM judges and preference evaluation.}
MT-Bench, Chatbot Arena, PandaLM, JudgeLM, G-Eval, Auto-J, Prometheus, and RewardBench have helped establish LLM judges as practical evaluators for open-ended generation \citep{zheng2023judging,wang2023pandalm,zhu2023judgelm,liu2023geval,li2023autoj,kim2024prometheus,kim2024prometheus2,lambert2024rewardbench}. These studies show that LLM judges can correlate with human preferences, but also that correlations depend on task, prompt, scoring scale, and model family. Our work follows this empirical tradition but treats the judge itself as the object of scaling analysis. The gap is that prior reports often present evaluator performance as a one-dimensional model ordering, leaving unclear whether gains arise from capability, benchmark slice, bias reduction, or protocol design.

\textbf{Judge bias and protocol effects.}
Prior work documents position bias, verbosity bias, sensitivity to rating format, and reliability changes caused by evaluation design choices \citep{dubois2024length,saito2023verbosity,wang2023fair,yamauchi2025designchoices}. These effects matter because they can change rankings without changing candidate responses. What remains under-specified is how bias changes jointly with evaluator scaling rather than under isolated prompt perturbations. Unlike a full prompt-factorial study, our experiments use a fixed main prompt and treat prompt sensitivity as a robustness check; the main focus is how model capability, bias, aggregation, and debate interact under a controlled reporting protocol.

\textbf{Aggregation and debate.}
Classical Condorcet-style results predict gains from independent voters, while work on correlated voters shows that dependence can sharply reduce those gains \citep{condorcet1785essai,ladha1992condorcet}. In LLM evaluation, self-consistency, multi-agent debate, and ChatEval-style protocols offer analogous routes to aggregation or deliberation \citep{wang2023selfconsistency,du2023debate,liang2023divergent,chan2023chateval,khan2024debating}. Existing studies often emphasize final accuracy; our focus is whether the aggregation assumptions required for such gains are empirically satisfied. Our jury experiment directly estimates error correlation, and our debate experiment is reported as an auditability case study.

\section{Experimental Design}

\subsection{Operationalizing Judge Reliability}

We treat LLM-as-judge reliability as a multi-dimensional measurement property rather than a single accuracy score. In this study, reliability consists of four observable components: judgment validity, bias robustness, aggregation independence, and protocol auditability. Table~\ref{tab:constructs} maps each construct to an operational measure and experiment.

\begin{table}[t]
\centering
\caption{Operationalizing LLM-as-judge reliability.}
\label{tab:constructs}
\scriptsize
\begin{tabular}{p{0.31\linewidth}p{0.46\linewidth}c}
\toprule
Construct & Operational measure & Exp. \\
\midrule
Judgment validity & Accuracy, Cohen's $\kappa$, Spearman & 1 \\
Bias robustness & Position flip, verbosity bias, granularity sensitivity & 2 \\
Aggregation independence & Error correlation $\rho$, jury gain & 3 \\
Protocol auditability & Parse status, fallback behavior, intermediate verdicts & 4 \\
\bottomrule
\end{tabular}
\end{table}

\subsection{Models and Datasets}

The main panel contains eight evaluated judges: Qwen3-1.7B, Qwen3-4B, Qwen3-14B, Qwen3-32B, MiniMax-M2, MiniMax-M2.1, MiniMax-M2.5, and MiniMax-M2.7. This panel separates an approximate parameter intervention from an observed released-model intervention. GLM-5.1 and mimo-v2-pro are used as cross-family reference judges in selected experiments, but are not members of either axis.

Four datasets cover complementary judgment settings so that reliability is not inferred from a single task slice. LLMBar is an adversarial pairwise benchmark with 419 examples \citep{zeng2024llmbar}. PandaLM testset-v1 provides broad-domain pairwise judgments; after removing tie-majority examples, 894 valid pairwise examples remain \citep{wang2023pandalm}. A 2,000-example seed-42 sample from Chatbot Arena provides broad human preference comparisons, with 1,997 valid pairwise examples \citep{zheng2023judging}. Our Judge's Verdict slice uses 200 TechQA-derived examples with three-level pointwise labels, supporting exact-match and rank-based evaluation \citep{han2025judgesverdict,castelli2019techqa}.

\subsection{Metrics and Protocols}

Pairwise experiments report accuracy and Cohen's $\kappa$ on parseable verdicts; Judge's Verdict additionally reports Spearman rank correlation. Main single-judge runs use near-greedy decoding ($T=0.1$) to reduce sampling noise when estimating model-level reliability. Homogeneous juries sample one judge multiple times at $T=0.7$ because repeated low-temperature calls would understate the dependence structure among stochastic jurors; heterogeneous juries combine different judges at $T=0.1$ to isolate model composition from sampling variance. Adjacent model comparisons use exact two-sided McNemar tests on parse-shared examples, so significance reflects paired judgment changes rather than differences in parser coverage. Holm correction controls the family of 18 adjacent tests.

The experiments instantiate the constructs in Table~\ref{tab:constructs}. Experiment 1 measures single-judge validity along the two axes. Experiment 2 measures LLMBar position, verbosity, and granularity robustness. Experiment 3 evaluates majority-vote juries under measured error correlation. Experiment 4 evaluates a structured debate protocol as an auditability case study. Four robustness checks cover human-ceiling calibration, corrected adjacent significance, limited prompt sensitivity, and Arena seed stability.

\begin{figure*}[t]
\centering
\includegraphics[width=0.92\textwidth]{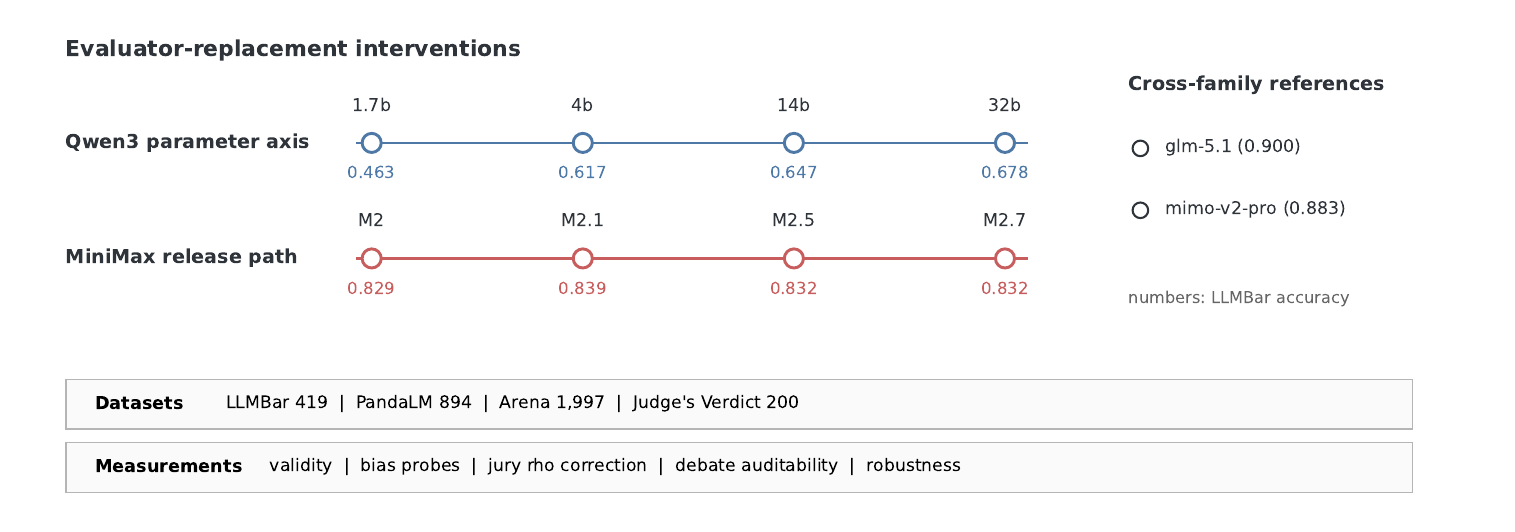}
\caption{Study design. The paper uses a Qwen3 parameter axis and a MiniMax release-generation axis as observable evaluator-replacement interventions, then tests whether single-judge scaling, bias probes, jury aggregation, and structured debate answer the same reliability question.}
\Description{A schematic of the two-axis LLM-as-judge testbed with model axes, datasets, experiments, and robustness checks.}
\label{fig:testbed}
\end{figure*}

\section{Results}

\begin{table*}[t]
\centering
\caption{Main answers by research question.}
\label{tab:rqanswers}
\scriptsize
\begin{tabular}{p{0.24\textwidth}p{0.18\textwidth}p{0.31\textwidth}p{0.19\textwidth}}
\toprule
Research question & Answer & Key evidence & Implication \\
\midrule
RQ1: Do scaling and release upgrades similarly improve reliability? & No. & Qwen3 1.7B$\rightarrow$4B is robustly significant; MiniMax adjacent releases are not significant. & Model upgrading is not a uniform reliability intervention. \\
RQ2: Does higher accuracy reduce bias? & Partly. & Position flips decline, but MiniMax-M2.7 still changes 14.7\% under A/B reversal. & High-accuracy judges still need bias probes. \\
RQ3: Do jury and debate protocols add reliability? & Conditionally. & Jury gains are small under high $\rho$; debate shifts are large but lack an audit trail. & Aggregation protocols must report dependence and audit trails. \\
\bottomrule
\end{tabular}
\end{table*}

\subsection{Finding 1: Judge Upgrades Are Not Interchangeable}

A capability-only account would predict similar adjacent gains along both upgrade paths; the paired tests do not show that pattern. Table~\ref{tab:singlejudge} summarizes the main single-judge pattern. Qwen3 improves sharply from 1.7B to 4B on LLMBar (0.463 to 0.617) and more modestly on Arena, but later parameter steps are smaller and not uniformly monotone. MiniMax release generations show no reliable adjacent improvement in this panel: adjacent pairwise differences are at most 0.022 accuracy, and none of the nine MiniMax adjacent tests reaches uncorrected $p<0.05$. After Holm correction across all 18 adjacent tests, only the Qwen3 1.7B to 4B steps on LLMBar and Arena remain significant.

\begin{table}[t]
\centering
\caption{Single-judge aggregate results. Values are accuracies except Judge's Verdict, where exact match is reported for comparability; brackets give Wilson 95\% intervals.}
\label{tab:singlejudge}
\scriptsize
\resizebox{\linewidth}{!}{%
\begin{tabular}{lcc}
\toprule
Dataset & Best judge & Qwen3 1.7B $\rightarrow$ 32B \\
\midrule
LLMBar & GLM-5.1, 0.900 [0.868,0.925] & 0.463 [0.416,0.511] $\rightarrow$ 0.678 [0.632,0.721] \\
PandaLM & MiniMax-M2.7, 0.857 [0.833,0.878] & 0.779 [0.751,0.805] $\rightarrow$ 0.769 [0.740,0.795] \\
Arena & mimo-v2-pro, 0.742 [0.722,0.761] & 0.625 [0.604,0.646] $\rightarrow$ 0.688 [0.667,0.708] \\
Judge's Verdict & GLM-5.1, 0.680 [0.612,0.741] & 0.595 [0.526,0.661] $\rightarrow$ 0.530 [0.461,0.598] \\
\bottomrule
\end{tabular}
}
\end{table}

The relevant conclusion is not that one family is universally stronger. No judge wins all datasets: GLM-5.1 leads LLMBar, MiniMax-M2.7 leads PandaLM, and mimo-v2-pro leads the sampled Arena slice. This shifts the design question from one-dimensional evaluator ordering to slice-specific measurement validity.

\begin{figure}[t]
\centering
\includegraphics[width=\linewidth]{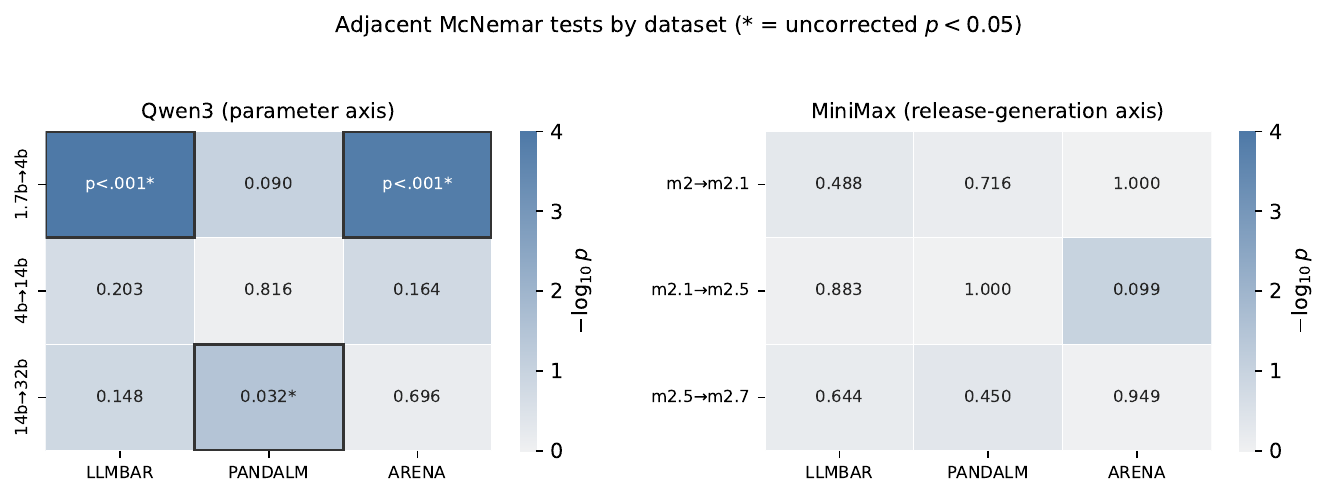}
\caption{Adjacent-pair McNemar tests for the two axes. After Holm correction over 18 adjacent tests, only the Qwen3 1.7B$\rightarrow$4B steps on LLMBar and Arena remain significant; no MiniMax adjacent release step is significant in this panel.}
\Description{A heat map of adjacent model significance values for Qwen3 and MiniMax on multiple datasets.}
\label{fig:mcnemar}
\end{figure}

\subsection{Finding 2: Stronger Judges Are Less Biased, Not Unbiased}

On LLMBar, the bias probes move in the same direction as capability, but they do not disappear. The position-flip rate falls from 0.320 for Qwen3-1.7B to 0.117--0.147 for MiniMax releases. Verbosity bias under a fixed generic and innocuous padding string falls from 0.547 for Qwen3-1.7B to roughly 0.13 for MiniMax. Granularity sensitivity is also smaller for the stronger judges.

This pattern is best interpreted as a capability--fairness association rather than a mechanism. Across the eight evaluated judges, the correlation between LLMBar accuracy and position-flip rate is strong (Pearson $r=-0.957$), but it is measured on one dataset and one model panel. Stronger judges are also not unbiased: MiniMax-M2.7 still changes 14.7\% of verdicts under A/B reversal. Thus position randomization and slice-level bias reporting remain necessary even when using high-accuracy judges.

\begin{figure}[t]
\centering
\includegraphics[width=\linewidth]{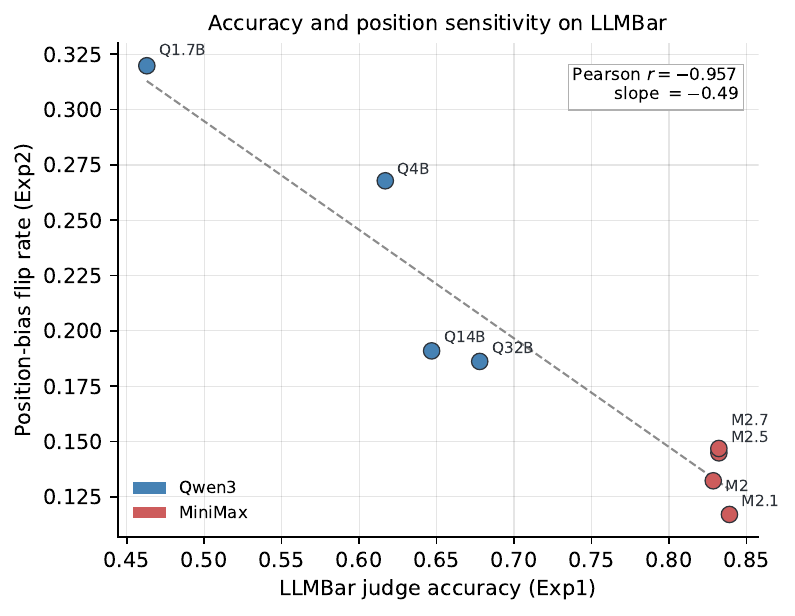}
\caption{Capability--fairness association on LLMBar. Higher single-judge accuracy co-varies with lower position-flip rate across the eight evaluated judges, but the non-zero flip rates show that stronger judges still require A/B randomization and bias reporting.}
\Description{A scatter plot of LLMBar accuracy versus position bias flip rate, with a negative linear trend.}
\label{fig:capfair}
\end{figure}

\subsection{Finding 3: Jury Size Matters Less Than Error Dependence}

The jury experiment fails for a familiar statistical reason: the votes are not independent. Majority voting is therefore not a generic reliability amplifier for LLM judges. For homogeneous juries, we estimate the intra-class error correlation $\rho$ from the vote matrix and compare three quantities: the independence prediction, a $\rho$-corrected beta-binomial prediction, and observed jury accuracy. Let $p$ be the single-judge baseline accuracy, $q\sim\mathrm{Beta}(\alpha,\beta)$, $\alpha=ps$, $\beta=(1-p)s$, and $s=1/\rho-1$. The jury accuracy is then the beta-binomial probability that a majority of $K$ votes are correct.

The correction is empirically necessary. Across homogeneous juries, the independence prediction has median absolute error 0.078 on LLMBar and 0.093 on PandaLM. The $\rho$-corrected prediction reduces these errors to 0.008 and 0.004, with maximum error below 0.02. Estimated correlations are high: Qwen3 homogeneous juries have $\rho=0.944$--0.972 on LLMBar, while MiniMax juries are lower but still correlated at $\rho=0.664$--0.706. As a result, increasing jury size barely moves accuracy; for Qwen3-1.7B on LLMBar, $K=1,3,5$ yields 0.463, 0.475, and 0.482.

Heterogeneous juries also underperform Poisson-binomial independence predictions. Family mixing alone therefore does not restore independent errors under a shared prompt. This shifts the relevant design question from ``how many jurors should be sampled?'' to ``how independent are their errors?''

\begin{figure}[t]
\centering
\includegraphics[width=\linewidth]{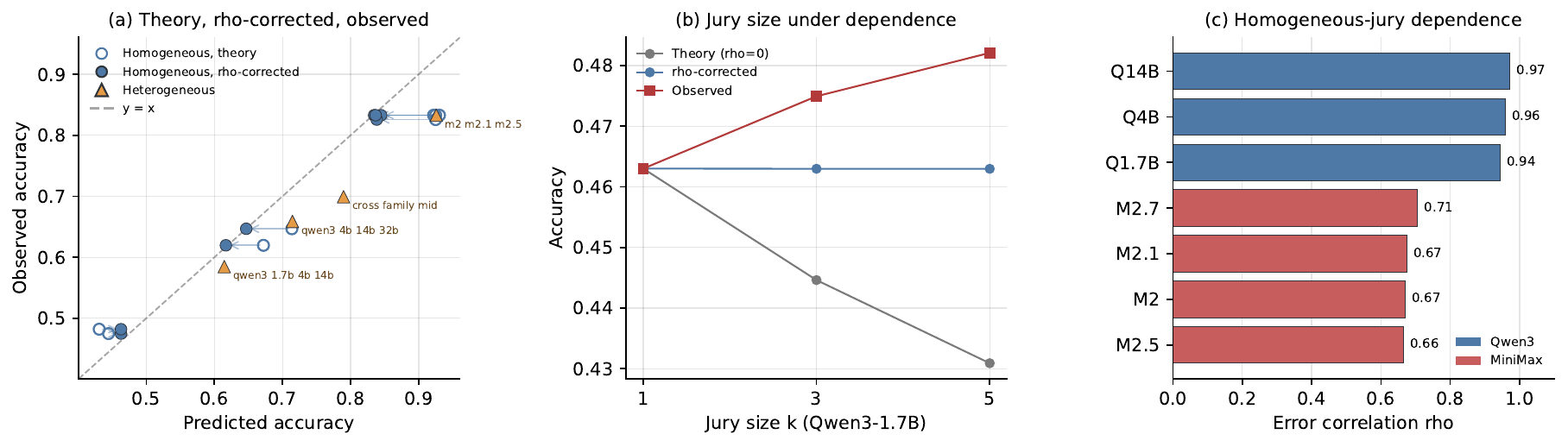}
\caption{Jury behavior on LLMBar. Majority voting provides little gain because sampled jurors make correlated errors; the $\rho$-corrected beta-binomial model tracks observed jury accuracy much better than the independence assumption.}
\Description{A multi-panel plot comparing independent, rho-corrected, and observed jury accuracy, plus jury-size and error-correlation panels.}
\label{fig:jury}
\end{figure}

\subsection{Finding 4: Debate Shifts Require Audit Trails}

The debate runs produce the largest protocol-level shifts in the paper, but they are also the least auditable. We include debate not as evidence for a deliberation effect, but as a stress test for protocol auditability. The structured-debate experiment pairs two judges for up to three rebuttal rounds on LLMBar. Cross-capability pairs show the largest final-vs-round-1 accuracy shifts: Qwen3-1.7B paired with GLM-5.1, MiniMax-M2.7, or mimo-v2-pro changes final decisions toward higher measured accuracy by $+0.317$, $+0.305$, and $+0.289$; Qwen3-4B paired with GLM-5.1 shifts decisions by $+0.243$ and reaches 0.897. Same-family and top-vs-top shifts are much smaller, ranging from $+0.005$ to $+0.050$.

These changes show that protocol-level interventions can alter judge outcomes substantially; the missing parser audit trail shows why such shifts should not be accepted without measurement logs. The implementation records round verdicts and final verdicts, but not raw responses or parse-success flags. Round-1 parse failures fall back to A, and later parse failures retain the previous verdict. Since fallback rates cannot be audited without rerunning the experiment, the debate results are best read as a suggestive protocol pattern and a reporting requirement, not as a clean estimate of deliberation.

\begin{figure}[t]
\centering
\includegraphics[width=\linewidth]{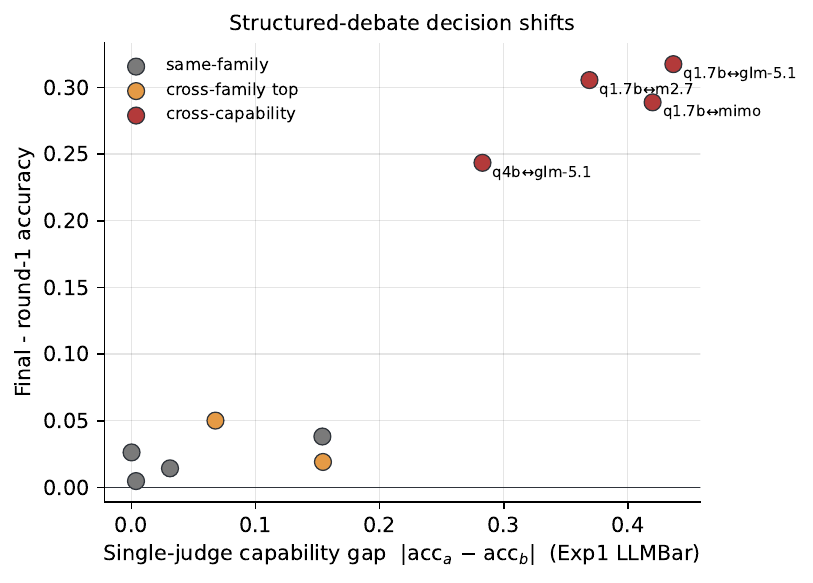}
\caption{Structured-debate accuracy shift versus single-judge capability gap on LLMBar. Cross-capability pairs produce the largest shifts, but the debate implementation's parse-fallback behavior makes the result suggestive rather than causal.}
\Description{A scatter plot showing debate accuracy shift increasing with single-judge capability gap.}
\label{fig:debategap}
\end{figure}

\subsection{Robustness Checks}

The robustness checks support the direction of the main findings while keeping their scope narrow. Human-ceiling calibration shows that PandaLM retains headroom: the best judge has matched leave-one-annotator-out $\kappa=0.753$ versus a human ceiling of 0.920. Judge's Verdict is closer to its noisy ceiling, with best-judge $\kappa=0.620$ versus human $\kappa=0.562$. Prompt-sensitivity checks on LLMBar preserve the cross-family ordering and show greater cross-prompt stability for stronger judges. Arena seed perturbations produce 0.004--0.032 accuracy spread, smaller than the robust Qwen3 1.7B to 4B Arena gap. These checks do not establish cross-family generality, but they reduce several obvious alternative explanations.

\section{Discussion}

\textbf{Principle.} A judge score is interpretable only when the evaluated slice, perturbation sensitivity, error dependence, and protocol state are jointly reported. The practical recommendation is therefore simple: an LLM-as-judge result should not be accepted as reliable unless it reports what slice was judged, how biased the judge was under simple perturbations, how correlated repeated judgments were, and whether protocol-level decisions were auditable.

\begin{table}[H]
\centering
\caption{A minimal audit trail for LLM-as-judge reports.}
\label{tab:reporting}
\tiny
\setlength{\tabcolsep}{2pt}
\renewcommand{\arraystretch}{0.92}
\begin{tabular}{p{0.27\linewidth}p{0.34\linewidth}p{0.29\linewidth}}
\toprule
Reliability risk & Required report item & Failure mode avoided \\
\midrule
Dataset-slice shift & Valid $N$, parseable $N$, tie handling & Hidden sample mismatch \\
Parser artifacts & Parser success rate, fallback rule & Artificial accuracy changes \\
Position/verbosity bias & A/B reversal, padding probe & Ranking artifacts \\
Correlated jury errors & $\rho$ and $K$, not $K$ alone & Overestimated voting gains \\
Debate protocol artifacts & Raw outputs, parser status, round verdicts & Unauditable protocol gains \\
Statistical overclaiming & Paired tests, Holm correction, uncertainty intervals where estimable & False adjacent-step claims or overstated precision \\
\bottomrule
\end{tabular}
\end{table}

The results suggest two methodological lessons for LLM-as-judge evaluation. First, judge scaling should be reported as a design space rather than a single evaluator ordering. In this panel, the clearest parameter-axis gain occurs at the low end, while the MiniMax release-generation axis changes little in the measured adjacent comparisons. This should not be read as a claim that all release generations lack meaningful gains, or that the MiniMax releases isolate one training stage. The supported conclusion is narrower: under the same datasets and prompts, the two observed upgrade paths have different adjacent-pair patterns.

Second, reliability reporting has to go beyond single-number accuracy and name the unit of analysis. Bias probes show that stronger judges can be less brittle, but they also show that bias remains measurable. Jury experiments show that aggregation gains depend on error correlation. Debate experiments show that protocol changes can be large enough to rival scaling effects, but only if the implementation preserves an audit trail for parse failures, tie-breaking, and abstentions. For pairwise datasets, this means parseable-subset definitions, parse-shared significance tests, and A/B randomization. For juries, it means reporting $\rho$ in addition to $K$. For debate or multi-agent protocols, it means storing raw responses, parser status, and every intermediate vote before interpreting final-score shifts. These requirements are operationally simple, but without them the apparent reliability of a judge can be an artifact of the evaluation pipeline.

\section{Threats to Validity}

\textbf{Construct validity.} Reliability is operationalized through selected proxies: accuracy and agreement, bias probes, error correlation, and auditability logs. These do not exhaust all reliability dimensions, such as calibration, long-form rationale quality, or domain-specific consistency.

\textbf{Internal validity.} The Qwen3 axis is closer to a parameter sweep, whereas the MiniMax axis is an observed release sequence rather than a controlled ablation. The debate experiment is especially limited by missing raw outputs and parse-success logs.

\textbf{Statistical conclusion validity.} McNemar tests are adjacent-pair tests and do not constitute a formal between-axis test. Multiple comparisons are controlled through Holm correction, but sample sizes and parser coverage differ across datasets.

\textbf{External validity.} Results are conditioned on two model families, two reference judges, four datasets, a small number of prompts, and one primary decoding regime per experiment. Additional domains, languages, and instruction distributions are needed before turning these observations into general judge-selection rules.

\section{Conclusion}

This paper presents a two-axis empirical study of LLM-as-judge reliability. The results show that evaluator capability is important but incomplete: reliability is also shaped by dataset slice, bias behavior, correlated errors, and protocol implementation. In the tested panel, the strongest robust adjacent effect is the Qwen3 1.7B to 4B transition; no MiniMax adjacent release step is significant; majority-vote juries are limited by high error correlation; and debate with a strong reference changes outcomes substantially but lacks sufficient evidence for causal interpretation under the current logs. These findings recast LLM-as-judge reliability as a measurement-validity problem rather than a model-selection problem alone.

\begin{acks}
This work was supported by the Open Research Project of the State Key Laboratory of Industrial Control Technology, China (Grant No. ICT2025B70); the Jiangxi Provincial Natural Science Foundation (Grant Nos. 20242BAB20041 and 20232BAB212006); the Hubei Provincial Natural Science Foundation of China (Grant Nos. 2023AFB474 and 2024AFB881); the Anhui Provincial Special Project for Special Needs in Humanities and Social Sciences (Grant No. 2025AHGXSK50067); and the Postgraduate Quality Engineering Project of Anhui Province (Grant No. 2024jyjxggyjY232).
\end{acks}

\bibliographystyle{ACM-Reference-Format}
\bibliography{references}

\end{document}